%% file: main.tex
\documentclass[10pt,twocolumn,letterpaper]{article}

\usepackage{cvpr}              


\definecolor{cvprblue}{rgb}{0.21,0.49,0.74}
\usepackage[pagebackref,breaklinks,colorlinks,allcolors=cvprblue]{hyperref}

\pdfoutput=1
\usepackage{enumitem}
\usepackage{graphicx}
\usepackage{multirow} 
\usepackage{threeparttable}
\usepackage{pifont}
\newcommand{\xmark}{\ding{55}}  
\usepackage{indentfirst} 
\newcommand{\dmark}{\scalebox{0.85}{\ding{52}}}

\usepackage[accsupp]{axessibility} 
\def\paperID{} 
\def\confName{CVPR}
\def\confYear{2026}

\author{Hui Xiang$^1$ 
Yifan Bian$^1$
Li Li$^1$
Jingran Wu
Xianguo Zhang
Dong Liu$^1$ \thanks{This work is supported by the Natural Science Foundation of China under Grant U25B2010. We acknowledge the support of GPU cluster built by MCC Lab of Information Science and Technology Institution, USTC. (Corresponding author: Dong Liu.)}  \\
$^1$ University of Science and Technology of China \\
{\tt\small xh0603@mail.ustc.edu.cn, togelbian@gmail.com, \{lil1, dongeliu\}@ustc.edu.cn}\\
}

\begin{document}
\input{sections/title}
\input{sections/abstract}
\input{sections/introduction}

\input{sections/related_work}

\input{sections/method}

\input{sections/experiments}

{
    \small
    \bibliographystyle{ieeenat_fullname}
    \bibliography{main}
}


\end{document}

%% file: sections/title.tex
\title{Real-Time Neural Video Compression with Unified Intra and Inter Coding}
\maketitle

%% file: sections/abstract.tex
\begin{abstract}
Neural video compression (NVC) technologies have advanced rapidly in recent years, yielding state-of-the-art schemes such as DCVC-RT that offer superior compression efficiency to H.266/VVC and real-time encoding/decoding capabilities. Nonetheless, existing NVC schemes have several limitations, including inefficiency in dealing with disocclusion and new content, interframe error propagation and accumulation, among others. To eliminate these limitations, we borrow the idea from classic video coding schemes, which allow intra coding within inter-coded frames. With the intra coding tool enabled, disocclusion and new content are properly handled, and interframe error propagation is naturally intercepted without the need for manual refresh mechanisms. We present an NVC framework with unified intra and inter coding, where every frame is processed by a single model that is trained to perform intra/inter coding adaptively. Moreover, we propose a simultaneous two-frame compression design to exploit interframe redundancy not only forwardly but also backwardly. Experimental results show that our scheme outperforms DCVC-RT by an average of 12.1\% BD-rate reduction, delivers more stable bitrate and quality per frame, and retains real-time encoding/decoding performances. The codes are available at \url{https://github.com/ihuixiang/UIIC}.

\end{abstract}

%% file: sections/introduction.tex
\section{Introduction}
Exploiting interframe redundancy lies at the core of video compression, as efficient removal of temporal redundancy directly translates to substantial bitrate savings. For decades, classic video coding standards \cite{h261,h.263,h.264,h.265,h.266} have continuously advanced temporal redundancy mining  via improved motion models \cite{ mesh1, wiegand2005affine, li2024uniformly} that enhance prediction accuracy. Recently, neural video compression (NVC) frameworks \cite{fvc,tcm, dc, tang2024offline, ssd,li2024neural, jiang2025ecvc, Bian_2025_CVPR, zhang2025flavc, tang2025neural, jia2025towards, tang2026neural} have further emphasized the utilization of inter-frame information to pursue higher compression efficiency. However, a critical oversight persists across most existing NVC designs: they prioritize exploiting inter-frame redundancy while neglecting to enhance intra-coding capabilities in scenarios where reference information is scarce or unreliable.

\begin{figure*}[ht]
\vspace{-4mm}
\centering
\includegraphics[width=\textwidth]{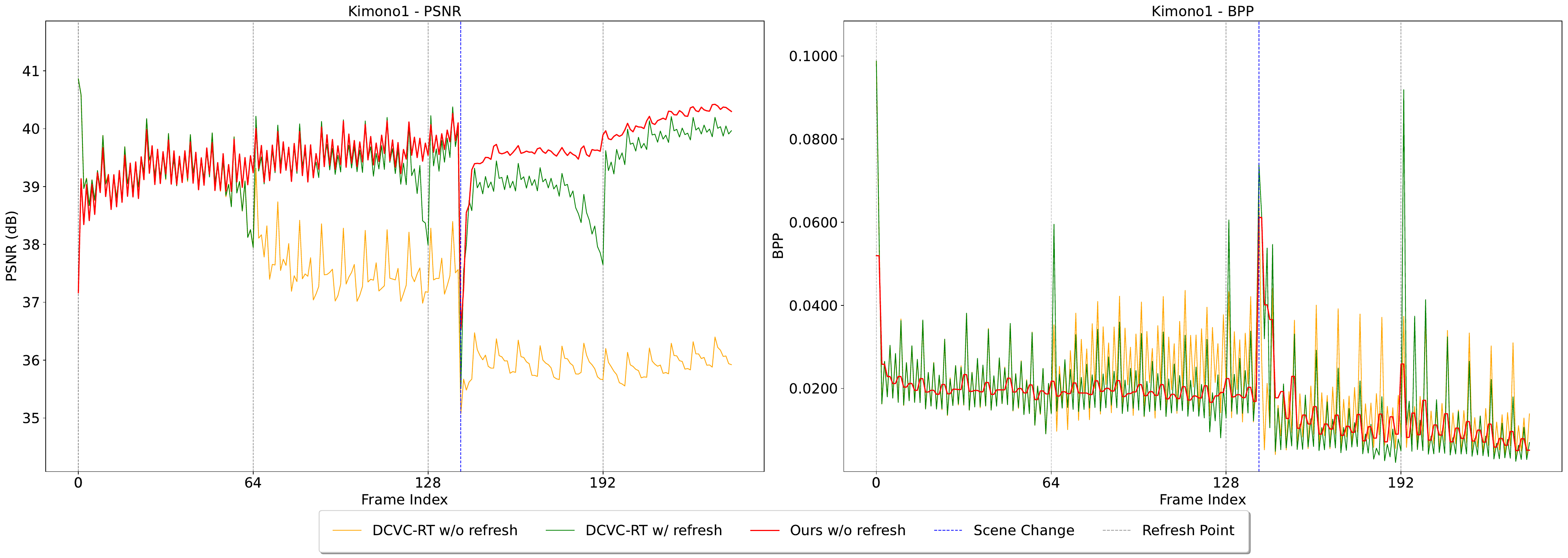}
\caption{Bitrate and quality variation across frames. The test video is Kimono1 from HEVC Class B, which contains a scene change at the 141st frame. Average bitrates in bpp: DCVC-RT (no refresh) 0.0203, DCVC-RT (with refresh) 0.0182, and ours 0.0172. Average PSNR in dB: DCVC-RT (no refresh) 37.33, DCVC-RT (with refresh) 39.33, and ours 39.57. Intra-period is set to -1; DCVC-RT (with refresh) uses a refresh period of 64; our scheme needs \emph{no} refresh mechanism.}
\vspace{-4mm}
\label{fig:kimono}
\end{figure*}

This limitation manifests acutely in practical scenarios such as scene changes. For instance, when the last frame of a preceding scene and the first frame of a new scene share no temporal correlation (Figure \ref{fig:kimono}), the P-frame model is forced to rely on its intrinsic intra-coding capacity. Given that state-of-the-art (SOTA) NVC schemes lack robust intra-coding support in P-frame designs, this mismatch leads to two key issues:  (1) significant quality degradation and (2) severe inter-frame error propagation—both of which compromise the quality of subsequent frames.

A second major challenge arises from GPU memory constraints during NVC training. The number of frames used for model training is typically insufficient to cover the full length of ultra-long sequences encountered in real-world testing. For such extended sequences, the heavy reliance of NVC schemes on previous frame features exacerbates error accumulation in reference signals. To mitigate this, periodic feature refresh techniques \cite{li2024neural} have been adopted in recent SOTA schemes \cite{jiang2025ecvc, Bian_2025_CVPR, tang2025neural, jia2025towards}: rich feature information is reconstructed into a three-channel pixel image via the recovery network, and this image is fed back as new reference features through an adaptor to intercept error accumulation. While this approach effectively blocks prior-frame errors, it suffers from two critical drawbacks: (1) it discards valuable inter-frame information (e.g., long-term temporal cues or occluded object details) alongside errors, and (2) it induces sharp bitrate surges at refresh points, posing risks of network congestion and hindering practical deployment. As shown in Figure \ref{fig:kimono}, the P-frames at refresh points exhibit bitrate patterns similar to intra-coded frames, yet existing models cannot autonomously correct propagation errors via adaptive intra-coding (due to weak P-frame intra capabilities), leaving them dependent on this inflexible refresh mechanism.


To address these core limitations, we propose a unified NVC framework that integrates intra- and inter-frame coding capabilities into a single model. Through training, we enable our model to adaptively balance between intra- and inter-frame coding. When reference information is accurate and abundant, the model prioritizes inter-frame prediction to maximize redundancy reduction; when references are error-prone or insufficient, it adaptively invokes intra-coding to enhance current-frame quality. This design naturally handles scene transitions, maintains stable bitrate and quality without manual refresh, and inherently mitigates inter-frame error propagation.

Balancing low bitrate, high visual quality, and real-time inference performance remains a critical and non-trivial challenge in reference-scarce scenarios (e.g., scene changes). Despite advancements in low-complexity real-time (RT) neural video compression (NVC) frameworks \cite{jia2025towards}, SOTA solutions still rely on computationally expensive standalone I-frame models to address these reference-scarce cases. Integrating such heavyweight intra-coding complexity directly into inter-frame coding pipelines would inevitably degrade inference speed, which is a key constraint for practical low-latency applications (e.g., real-time video streaming). To resolve this fundamental trade-off, we propose a simultaneous two-frame compression technique: by jointly encoding two consecutive frames, this approach leverages rich backward reference information from the subsequent frame, while incurring merely one frame latency. Moreover, the collaborative modeling of inter-frame dependencies enables the extraction of more comprehensive temporal cues (e.g., fine-grained temporal trends and cross-frame object correlations) that are inaccessible in single-frame coding. 

Our key contributions are summarized as follows:
\begin{itemize}[leftmargin=10pt, nosep]
\item We unify intra- and inter-frame coding into a single model, eliminating the need for a separate intra-coding model. This design enhances the NVC’s ability to handle scene changes and reduces model parameter count.
\item We train the unified model to adaptively balance intra- and inter-coding based on reference quality, directly addressing inter-frame error propagation and bitrate spikes caused by manual refresh mechanisms.
\item We propose a simultaneous two-frame compression technique that leverages backward references from subsequent frames, preserving real-time inference speed while maximizing inter-frame redundancy exploitation.
\item Experimental results demonstrate our framework outperforms the SOTA low-complexity NVC scheme, i.e., DCVC-RT \cite{jia2025towards} in the most challenging configuration by an average of 12.1\% BD-rate reduction, with a smaller model size and comparable inference speed.
\end{itemize}

%% file: sections/related_work.tex
\begin{figure*}[t]
  \vspace{-2mm}
  \centering
  \includegraphics[width=\textwidth]{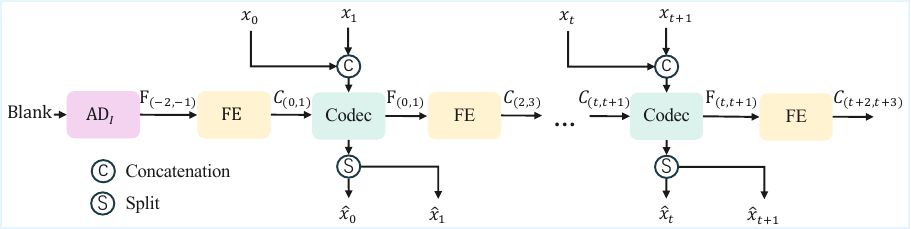}
  \caption{Framework of our neural video compression model with unified intra and inter coding. $x_0,x_1,x_t,x_{t+1}$ denote the first, second, $(t+1)$-th, $(t+2)$-th frames of the video sequence, and $\hat{x}_0,\dots$ are the corresponding reconstructed frames, respectively. $AD_I$ is the first-frame-referenced adaptor that accepts a blank signal as input, $FE$ stands for the feature extractor, and Codec denotes the encoder and decoder network. $F_{(t, t+1)}$ represents the intermediate features generated during the encoding/decoding of $x_t$ and $x_{t+1}$, while $C_{(t, t+1)}$ denotes the reference features for $x_t$ and $x_{t+1}$.}
  \label{fig:intra-inter}
  \vspace{-4mm}
\end{figure*}

\section{Related Work}
Classic video coding standards (e.g., H.264/AVC, H.265/HEVC, H.266/VVC) \cite{h261,h.263,h.264,h.265,h.266} have long relied on inter-frame prediction to reduce temporal redundancy, where the current frame is predicted using reference frames via motion estimation and compensation. A critical design in these standards is the integration of intra prediction within inter-frame coding: when inter prediction fails to yield satisfactory performance (e.g., for disoccluded regions, newly appearing content, or areas with complex motion where motion vectors cannot capture accurate dependencies), local blocks are switched to intra coding mode. This strategy avoids excessive residual distortion caused by poor inter prediction, balancing compression efficiency and reconstruction quality. The same principle retains significant value for modern neural video coding (NVC) frameworks, where addressing scene inhomogeneity and inter-frame error propagation remains a fundamental challenge.

Recent advances in NVC have demonstrated great potential in video compression. Existing schemes can be broadly categorized into two main types: residual coding and conditional coding. Residual coding~\cite{res_1, res_2, res_3, res_4, dvc, dvc_pro, fvc, hu2022coarse, agustsson2020scale} primarily reduces redundancy by encoding the difference between predicted frames and the current frame. Conditional coding~\cite{cond_1, cond_2, dcvc, tcm, ho2022canf, hem, dc, li2024neural, jiang2025ecvc, Bian_2025_CVPR, zhang2025flavc, tang2025neural, jia2025towards} focuses on exploiting contextual correlations to improve coding performance. Some other emerging frameworks based on neural video representations~\cite{chen2021nerv,chen2023hnerv,zhao2023dnerv,li2022nerv, nvrc2024, gao2025givic, liu2025an} are also rapidly developing. Given that conditional coding can learn complex temporal patterns and has entropy lower than or equal to that of residual coding, real-time conditional coding is increasingly approaching practical applicability with the advent of DCVC RT~\cite{jia2025towards}. Our model adopts a real-time conditional coding design and introduces improvements in reducing inter-frame error propagation.

Conditional coding leverages contextual information for prediction, where the prediction performance depends on the richness and accuracy of the context. Most existing NVC schemes enhance the context by efficiently utilizing temporal references~\cite{jiang2025ecvc, Bian_2025_CVPR, zhang2025flavc, tang2025neural, jia2025towards}. For instance, 
DCVC-FM~\cite{li2024neural} proposes training with longer sequences and a long-sequence refresh mechanism to maintain the accuracy of the temporal context. ECVC~\cite{jiang2025ecvc} utilizes non-local temporal context to enrich the reference information for the current frame. 
SEVC~\cite{Bian_2025_CVPR} preserves richer contextual information through multi-resolution temporal contexts. However, these methods rely on optical flow networks, which often entail high computational complexity and are unsuitable for real-time applications. 
In contrast, DCVC-RT~\cite{jia2025towards} explores temporal context correlations via implicit context alignment, eliminating the need to encode additional motion information. This approach significantly reduces computational complexity, making the practical deployment of NVC feasible.
Nevertheless, DCVC-RT still faces unresolved critical challenges, including inefficiency in handling disocclusion and new content, as well as inter-frame error propagation and accumulation, which continue to restrict its further practical deployment. This technical gap serves as the key motivation for the core innovation of our work.

%% file: sections/method.tex
\begin{figure*}[ht]
  \centering
  \includegraphics[width=\textwidth]{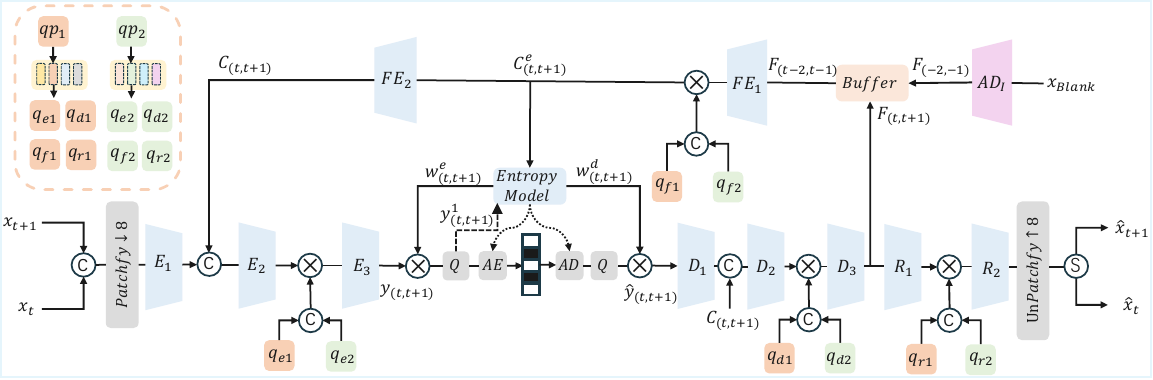}
  \caption{Detailed architecture of our proposed UI$^2$C (unified intra and inter coding) scheme. $(E_1, E_2, E_3)$, $(D_1, D_2, D_3)$, $(R_1, R_2)$, $(FE_1, FE_2)$ represent the components of the encoder, decoder, reconstruction generator, and feature extractor, respectively (detailed configurations are provided in the supplementary material).
  The reference feature $F_{(t-2, t-1)}$ is used to obtain the contextual feature $C^e_{(t, t+1)}$ and $C_{(t, t+1)}$. The quantization parameters $qp_1$ and $qp_2$ are used to select the respective quantization tables. $\times$ stands for element-wise multiply.}
  \vspace{-4mm}
  \label{fig: overview-detail}
\end{figure*}

\section{Method}
\subsection{Overview}
Our proposed model, namely UI$^2$C (\textbf{}{U}nified \textbf{I}ntra and \textbf{I}nter \textbf{C}oding), builds on the real-time neural codec DCVC-RT \cite{jia2025towards}, with core enhancements to address prior limitations. As shown in Figure~\ref{fig:intra-inter}, UI$^2$C removes dedicated I-frame models by integrating unified intra-inter coding into a single spatio-temporal network (Section \ref{unified intra- and inter- frame coding}).
To encode \(x_t\), when $t$ is even ($t=0, 2,\dots$), 1-frame latency is introduced to wait for \(x_{t+1}\) for simultaneous two-frame compression. These two frames are concatenated and fed to a shared encoder-decoder, exploiting forward (prior decoded frames) and backward (\(x_{t+1}\) for $x_t$) temporal redundancies \cite{jia2025towards}. The decoder reconstructs both frames from one piece of code, retaining their features in the reference buffer for subsequent coding (Section \ref{one-frame lookahead}).

To optimize rate-distortion (RD) performance across the two frames, a two-frame quantization table strategy allocates bitrates by frame indexes (Section \ref{dual quantization}). For training, a hybrid reference scheme enhances adaptive coding: initial batch frames randomly use blank references (simulating intra-dominant scenarios) or noise-perturbed prior features (mimicking error-prone inter references), enabling UI$^2$C to switch modes based on reference reliability (Section \ref{hybrid reference}).

\subsection{Unified intra- and inter-frame coding}
\label{unified intra- and inter- frame coding}
In previous NVC schemes \cite{dvc, dcvc, li2024neural, jia2025towards}, divided models have been employed for I-frames and P-frames, allowing the I-frame model to fully specialize in intra-coding capabilities and the P-frame model to excel in inter-coding. 

Most existing NVCs utilize an I-frame model to process the first frame, where intra-coding occurs without any reference. A similar situation arises during scene changes, where there is no correlation between previous frames and the current frame. In such cases, the P-frame model is forced to employ intra-coding. These two scenarios are essentially identical: both involve encoding the current frame without any available reference. 

Moreover, since the number of available P-frames for training is inherently limited, existing models are unable to automatically handle extremely long sequences without external manual intervention, such as inserting I-frames or applying periodic refresh mechanisms. Invoking the I-frame model truncates all potential errors through intra-coding, which discards all potentially useful references. This leads to the RD performance of IP32 (Intra Period=32 frames) configurations consistently underperforming compared to IP-1 (One Intra frame). The manual refresh method \cite{li2024neural} converts accumulated features back to the pixel-domain image, which is then processed by an adaptor to generate new features. This approach preserves partial information (only spatial information from the previous frame, while all temporal information is discarded) and simultaneously eliminates errors. However, this forces the subsequent first P-frame to encode with limited information, making the model treat it as the first P-frame after an I-frame, thereby initiating a new coding cycle. This lack of information compels the P-frame model to resort to more intra-coding. Yet, the current P-frame model exhibits weak intra-coding capabilities and struggles to effectively balance intra- and inter-coding, resulting in excessively high bitrates for that frame.

Based on this observation, we argue that a dedicated I-frame model is unnecessary. Instead, a single unified model can effectively handle both intra- and inter-coding scenarios. During inference, for the first frame, we feed a blank frame through an adaptor to generate reference features, enabling the model to directly use its inherent intra-coding capability. As illustrated in Figure~\ref{fig:intra-inter}, the model processes the first frame by converting the blank input into reference features via the adaptor, thereby activating its intra-coding behavior. For subsequent frames, the same model is reused with informative reference features, allowing it to primarily exploit its inter-coding capabilities.

\subsection{Simultaneous two-frame compression}
\label{one-frame lookahead}

In practical low-latency scenarios (e.g., real-time streaming), 1-frame latency is widely acceptable with high frame rate\cite{jia2025towards}, creating opportunities to leverage backward references from the subsequent frame. This bidirectional temporal redundancy exploitation is critical for resolving the key trade-off in our unified framework: maintaining low complexity while enhancing coding robustness.

In reference-scarce cases (e.g., the first frame or scene transitions), backward references from $x_{t+1}$ compensate for the lack of preceding frame information, alleviating the quality degradation that would otherwise arise from weak intra-coding under constrained complexity. For inter-coding, the bidirectional cues enable more accurate modeling of occluded regions and offer error calibration for noisy or imperfectly propagated features.

Notably, consecutive frames in natural video sequences exhibit substantial temporal redundancy with high similarity. After 8× joint downsampling, trivial high-frequency variations between the two frames are suppressed, further enhancing their feature-level consistency to enable efficient joint encoding. As illustrated in Figure~\ref{fig: overview-detail}, building on DCVC-RT’s efficiency-driven design \cite{jia2025towards}, we concatenate $x_t$ and $x_{t+1}$ along the channel dimension, apply the aforementioned joint downsampling, and feed the fused feature into the shared encoder-decoder. This single-stream pipeline leverages the implicit modeling capability of neural networks for mutual reference between the two frames, generating only one compact bitstream while preserving real-time inference speed. At the decoder side, both frames are reconstructed synchronously; the fused features generated here are stored in the reference buffer, providing richer contextual information for the coding of subsequent frames and ultimately forming a performance-enhancing loop.

\subsection{Two-frame quantization}
\label{dual quantization}
Joint two-frame compression introduces a critical RD optimization challenge: retaining the efficiency of the Hierarchical Quality Structure \cite{dc} while enabling fine-grained quality control between co-encoded frames. Existing real-time neural codecs like DCVC-RT \cite{jia2025towards} rely on shared quantization tables for rate control across encoders, decoders, reconstruction generators, and feature extractors—but this design fails to account for the distinct reference roles of the two frames (e.g., \(x_{t+1}\) serves as both backward reference for \(x_t\) and future reference for subsequent frames, while \(x_t\) focuses on forward context).

To regulate the quality control between the two frames during encoding, we adopt a \textit{two-frame quantization} approach. As is shown in Figure~\ref{fig: overview-detail}, for the two input frames, each frame is assigned a quality parameter (\textit{qp}) queried based on its frame index, resulting in two distinct \textit{qp} values. For these two \textit{qp} values, corresponding quantization coefficients are obtained by querying different quantization tables. The quantization coefficients of both frames are then concatenated and multiplied by the features at corresponding positions to achieve quality control. Moreover, we assign a higher \textit{qp} to the latter frame of the two frames, so that subsequent frames have a better reference.

\subsection{Training with hybrid references}
\label{hybrid reference}
With the unified intra-inter coding model in place, designing an effective training strategy to unlock its full performance becomes critical. While prior works \cite{li2024neural, Bian_2025_CVPR, jia2025towards} have released their testing code, their training methodologies remain non-trivial to replicate. Though \cite{li2025opendcvcs} explored training strategies for the DCVC series, its final performance still lags behind officially released models.

The core challenge in training UI$^2$C is enabling the model to dynamically balance intra- and inter-coding based on the current reference error level. For the reference of initial frames, we consider three candidates: a pure blank signal (e.g., an all-zero image), the ground-truth (GT) of the previous frame, and a noise-corrupted version of this GT. 
For the noisy-reference features, we infer them from previously reserved frames and use the resulting features as training references.
During training, we randomly sample one of these three as the initial frame’s reference. This strategy forces the model to learn to implicitly assess reference error levels, allowing it to adaptively enhance intra-coding for error correction when processing sequences longer than the training data, without requiring manual reference discarding. 
Eliminating the need for info-discarding refresh mechanisms also reduces peak bitrate, thereby mitigating the risk of network congestion.

%% file: sections/experiments.tex
\begin{table*}[t]
\centering
\vspace{-4mm}

\caption{BD-rate (\%) with DCVC-RT as the anchor and encoding/decoding speeds. Color space is YUV420; all frames are coded; intra-period is –1.}
\label{tab:results}
\setlength{\tabcolsep}{7pt} 
\begin{tabular}{@{}lccccccccc@{}}
\toprule
 & \multicolumn{7}{c}{BD-Rate (\%)} & \multicolumn{2}{c}{Speed (fps)} \\
\cmidrule(lr){2-8} \cmidrule(lr){9-10}
Method & HEVC B & HEVC C & HEVC D & HEVC E & MCL-JCV & UVG & Average & Enc. & Dec. \\
\midrule
VTM-17.0 &15.7 & 21.1 & 34.7 & 28.0 & 13.8 & 28.5 & 23.6 &0.01&20.5\\
DCVC-DC & 22.1 & -0.5 & -5.7 & 145.2 & 4.4 & 33.1 & 33.1 & 1.6 & 1.9  \\
DCVC-FM & -1.4 & -13.9 & -16.9 & -7.7 & 4.5 & 3.9 & -5.3 & 1.5  & 1.7  \\
DCVC-RT & 0.0 & 0.0 & 0.0 & 0.0 & \textbf{0.0} & 0.0 & 0.0 & 56.8  & \textbf{51.5}  \\
UI$^2$C (Ours) & \textbf{-9.8} & \textbf{-16.4} & \textbf{-23.5} & \textbf{-17.7} & 1.1 & \textbf{-6.1} & \textbf{-12.1} & \textbf{65.1}  & 46.1  \\
\bottomrule
\vspace{-4mm}
\end{tabular}
\end{table*}

\begin{figure*}[tbp]
\vspace{-4mm}
\centering
\begin{minipage}[b]{0.32\linewidth}
    \centering
    \includegraphics[width=\linewidth]{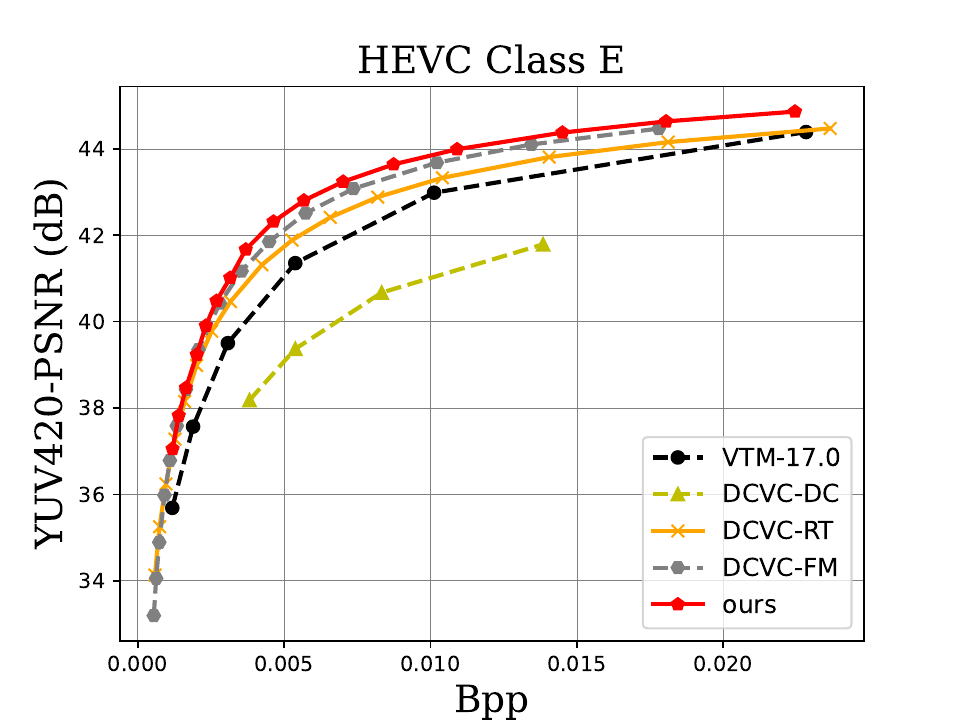}
\end{minipage}
\hfill
\begin{minipage}[b]{0.32\linewidth}
    \centering
    \includegraphics[width=\linewidth]{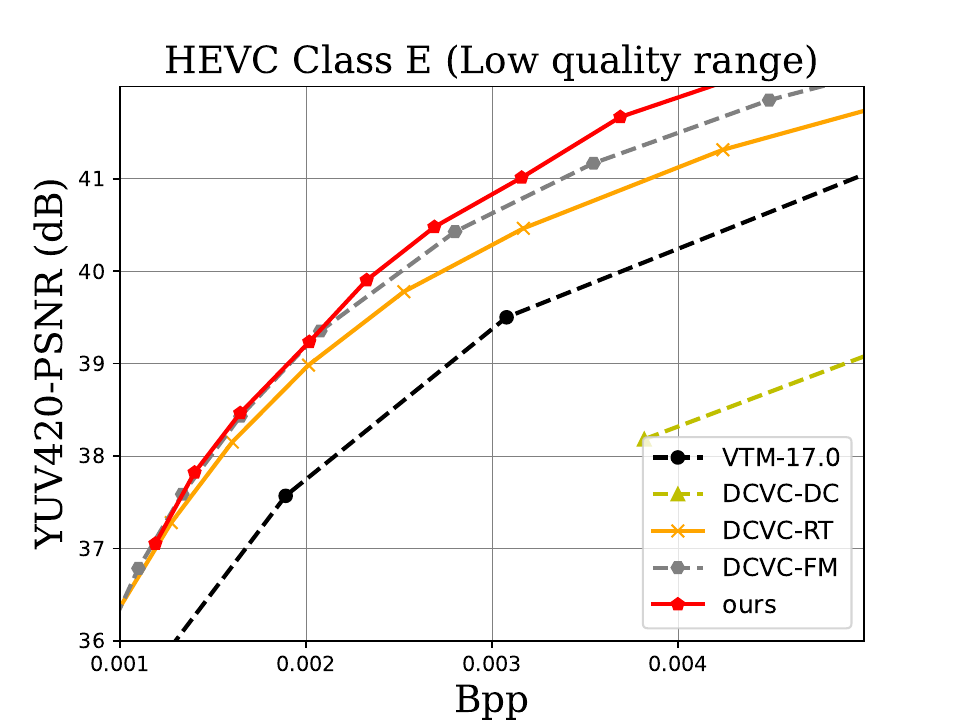}
\end{minipage}
\hfill
\begin{minipage}[b]{0.32\linewidth}
    \centering
    \includegraphics[width=\linewidth]{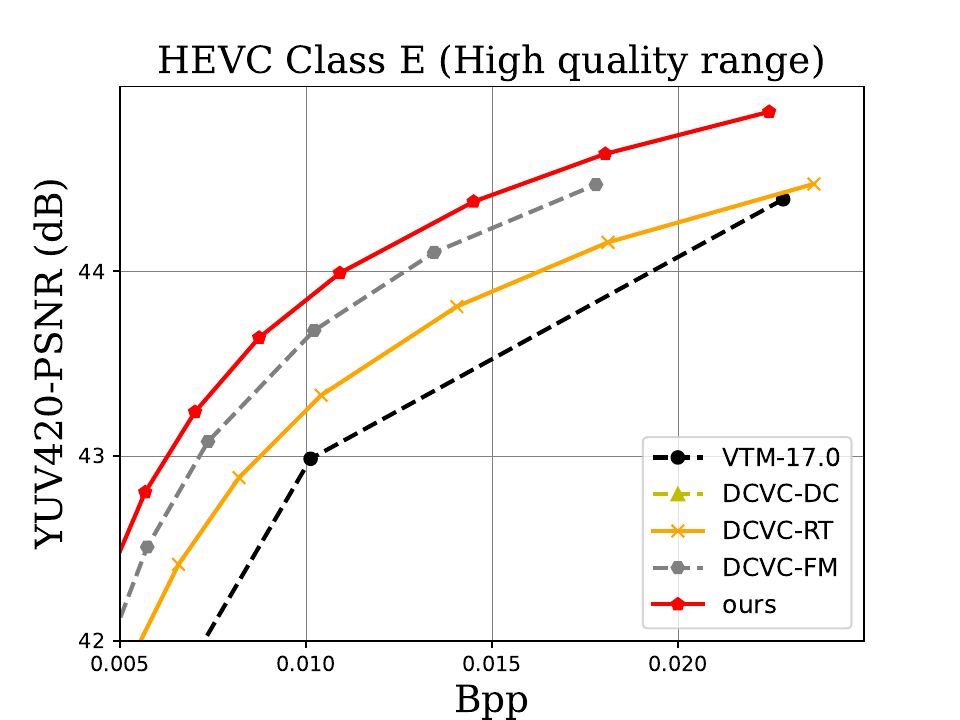}
\end{minipage}

\begin{minipage}[b]{0.32\linewidth}
    \centering
    \includegraphics[width=\linewidth]{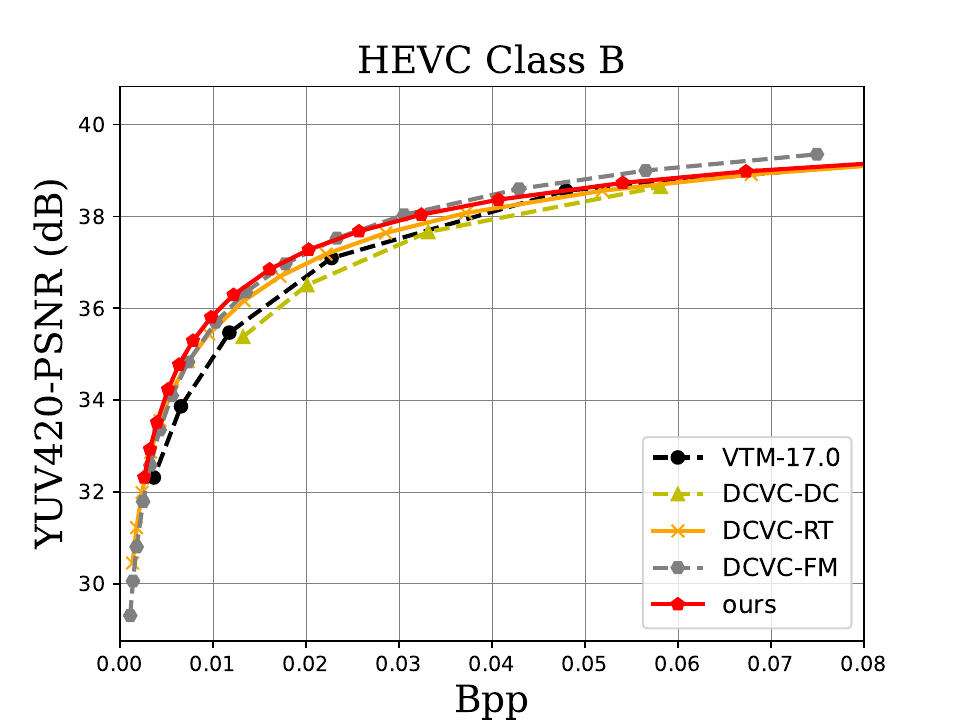}
\end{minipage}
\hfill
\begin{minipage}[b]{0.32\linewidth}
    \centering
    \includegraphics[width=\linewidth]{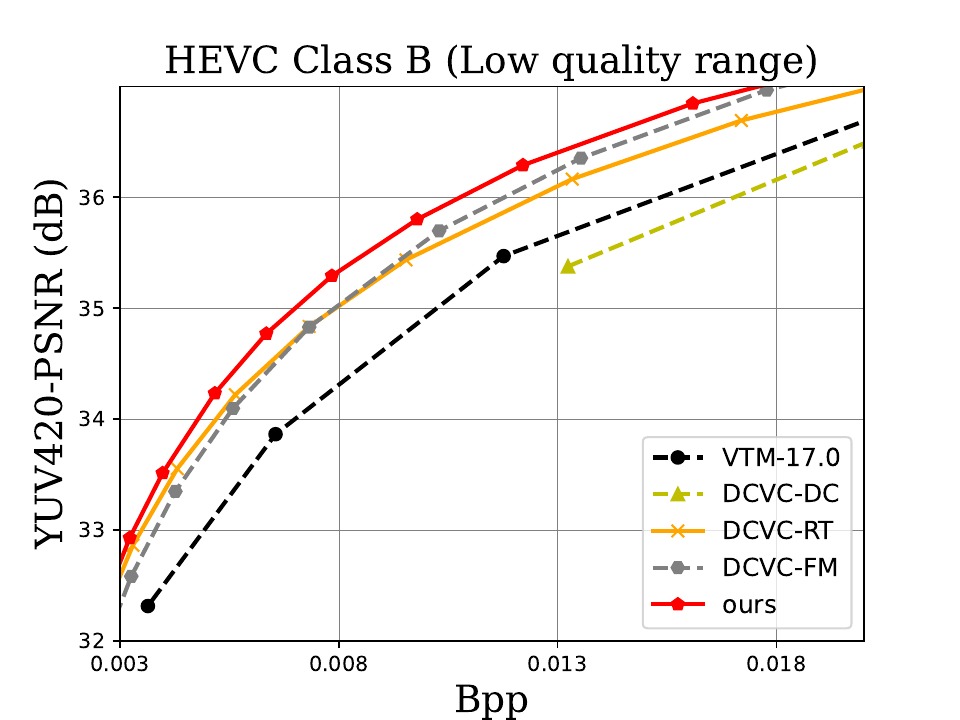}
\end{minipage}
\hfill
\begin{minipage}[b]{0.32\linewidth}
    \centering
    \includegraphics[width=\linewidth]{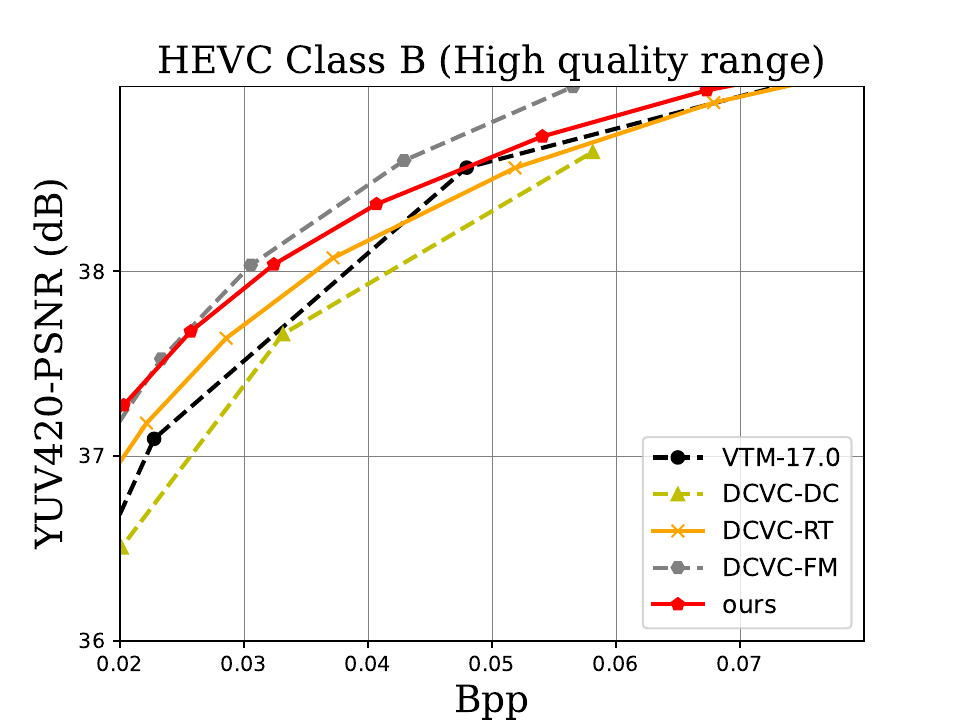}
\end{minipage}

\caption{Rate-distortion curves for HEVC Class E and HEVC Class B. Color space is YUV420; all frames are coded; intra-period is -1. Results on more datasets are provided in the supplementary material.}
\vspace{-4mm}
\label{fig:rd-curve}
\end{figure*}


\begin{figure*}[t]
\centering
\vspace{-4mm}
\begin{minipage}[b]{0.32\linewidth}
    \centering
    \includegraphics[width=\linewidth]{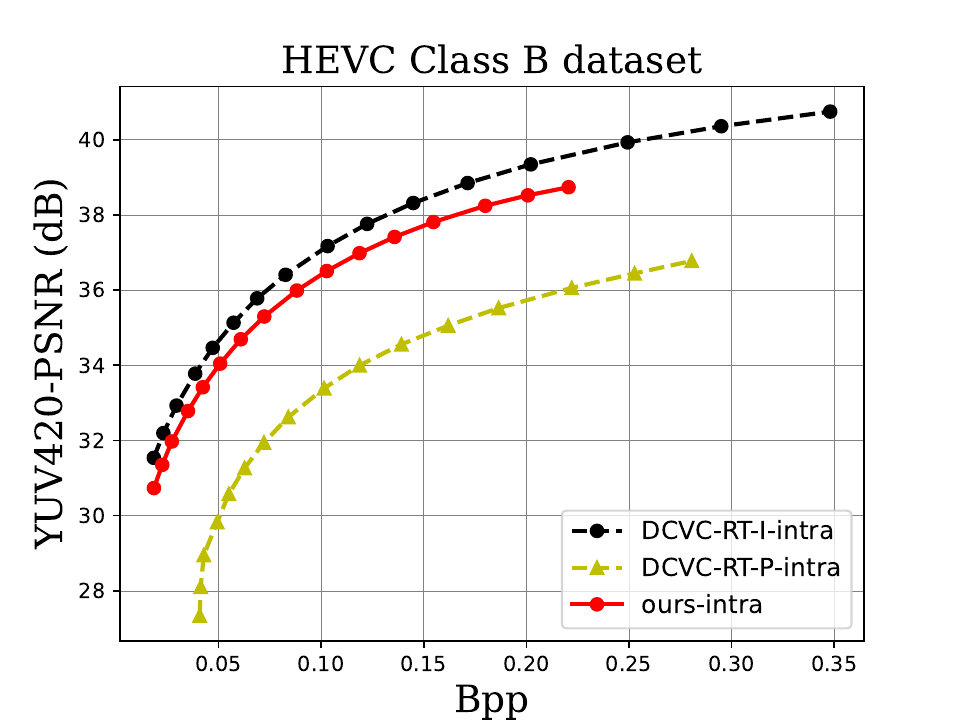}
\end{minipage}
\hfill
\begin{minipage}[b]{0.32\linewidth}
    \centering
    \includegraphics[width=\linewidth]{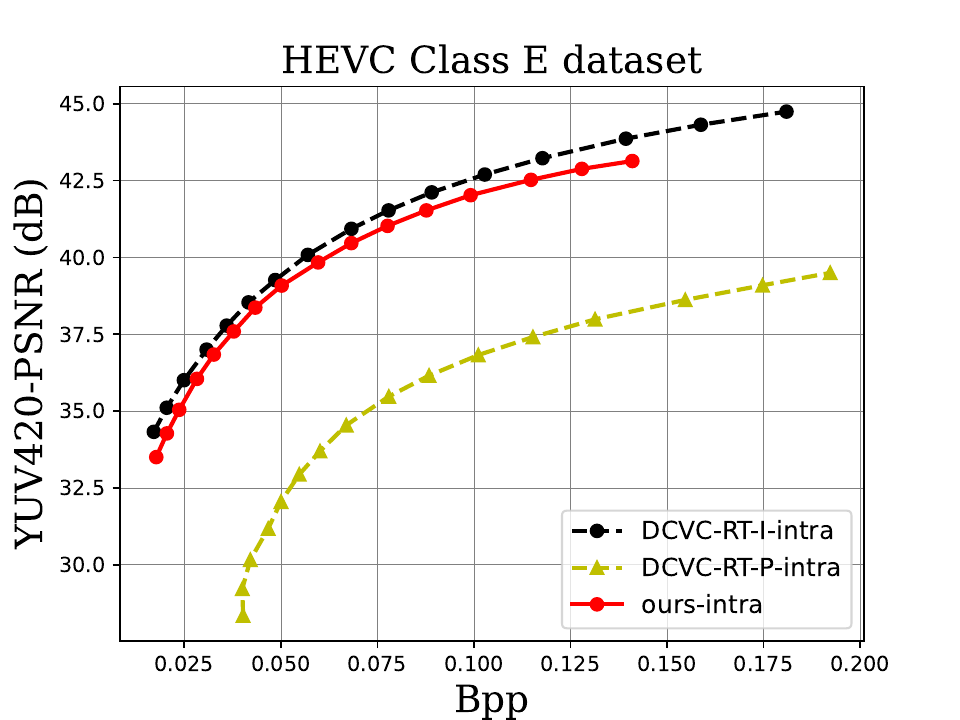}
\end{minipage}
\hfill
\begin{minipage}[b]{0.32\linewidth}
    \centering
    \includegraphics[width=\linewidth]{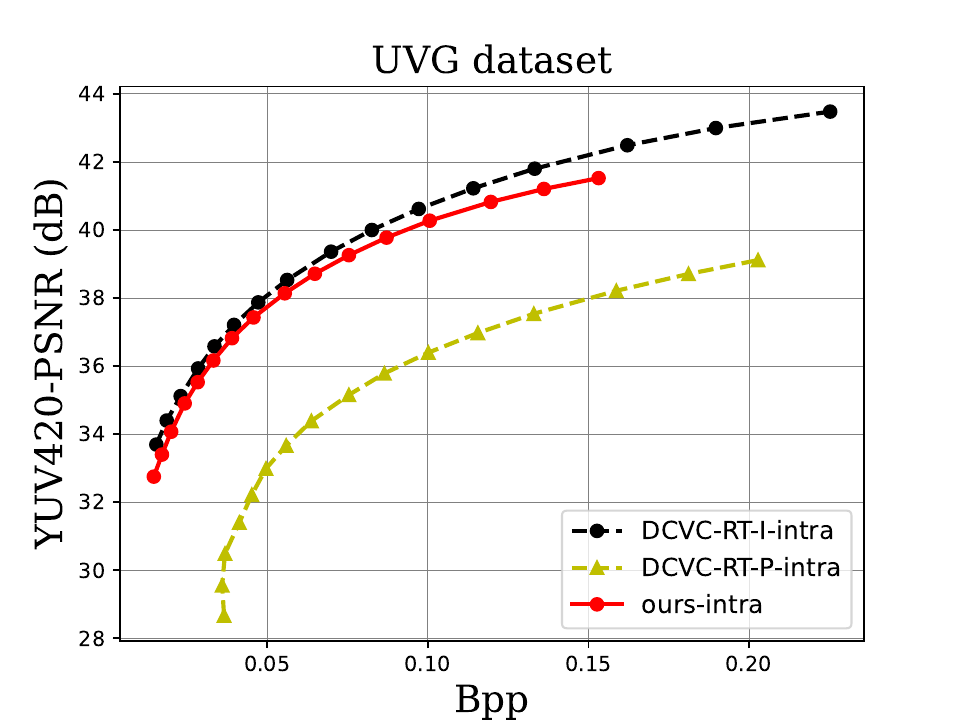}
\end{minipage}
\caption{Rate-distortion curves for HEVC Class B, HEVC Class E, and UVG when \emph{only the first two frames} are coded. DCVC-RT-I-intra uses the normal DCVC-RT setting (the first frame uses the intra coding model, and the second frame uses the inter coding model). DCVC-RT-P-intra \emph{only} uses the DCVC-RT inter coding model, i.e., we insert a blank frame (all zeros) before the first frame, and do not count the bitrate/PSNR of the blank frame. Our method using a single model to compress the two frames simultaneously performs slightly worse than DCVC-RT-I-intra, which uses two models for intra and inter coding, respectively, but performs much better than DCVC-RT-P-intra. }
\label{fig:rdintra}
\vspace{-2mm}
\end{figure*}

\begin{table}[t]
\centering
\small
\setlength{\tabcolsep}{1pt} 
\begin{threeparttable}
\caption{Computational complexity comparison}
\label{tab:complexity}
\begin{tabular}{@{}lccccc@{}}
\toprule
Model & \begin{tabular}[c]{@{}c@{}}Encoding\\ (kMACs/pixel)\end{tabular} & \begin{tabular}[c]{@{}c@{}}Decoding\\ (kMACs/pixel)\end{tabular} & \begin{tabular}[c]{@{}c@{}}Params.\\ (M)\tnote{*}\end{tabular} & \begin{tabular}[c]{@{}c@{}}Latent\\ Channels\tnote{\dag}\end{tabular} & \begin{tabular}[c]{@{}c@{}}Dec.\\ Steps\tnote{\dag}\end{tabular}\\ 
\midrule
DCVC-DC & 1333 & 910 & 50.9 & 128 & 4  \\
DCVC-FM & 1137 & 866 & 45.0 & 128 & 4   \\
DCVC-RT & 142 & 167 & 66.4 & 128 & 2  \\
UI$^2$C (Ours) & 157 & 233 & 46.7 & 64 & 1  \\
\bottomrule
\end{tabular}

\begin{tablenotes}
    \footnotesize
    \item[*] Params: Total parameter count of both I-frame and P-frame models; ours has only one model.
    \item[\dag] Latent channels and Dec. steps are average values per frame. Dec. step refers to autoregressive steps in the entropy model (excluding optical flow and hyperprior).
  \end{tablenotes}
\end{threeparttable}
\end{table}



\begin{table*}[t]
\centering
\setlength{\tabcolsep}{6pt} 
\caption{Ablation studies for BD-rate with the full model without refresh as the anchor}
\begin{tabular}{@{}cccc|ccccc@{}}
\toprule
\multicolumn{4}{c|}{Configurations} & \multicolumn{5}{c}{BD-rate (\%)} \\
\cmidrule(lr){1-4} \cmidrule(lr){5-9}
Unified & Two-frame Compr. & Hybrid Ref. & Refresh& HEVC-B & HEVC-C & HEVC-D & HEVC-E & Average \\
\midrule
\xmark & \xmark & \xmark & 64 & 25.7 & 26.2 & 36.7 & 46.6  & 33.8\\
\dmark & \xmark  & \xmark & 64 & 43.9 & 44.3 & 60.1 & 117.3 & 66.4\\
\dmark & \dmark  & \xmark & 64 & 23.5 & 22.7 & 26.2 & 55.6 & 32.0\\
\dmark  & \dmark  & \dmark & 64 & 24.4&21.5&27.2&61.6 & 33.7 \\
\xmark & \xmark & \xmark & -- & 63.3 & 61.8 & 71.7 & 178.6 & 93.9\\
\dmark & \xmark  & \xmark & -- &  18.3 & 21.2 & 31.2 & 45.3 & 29.0\\
\dmark & \dmark  & \xmark & -- & 5.1 & 3.2 & 4.3 & 8.5 & 5.3  \\
\dmark  & \dmark  & \dmark & -- & 0.0 & 0.0 & 0.0 & 0.0 & 0.0 \\
\bottomrule
\end{tabular}
\label{tab:ablation}
\end{table*}

\begin{figure*}[tbp]
  \centering
  \includegraphics[width=\textwidth]{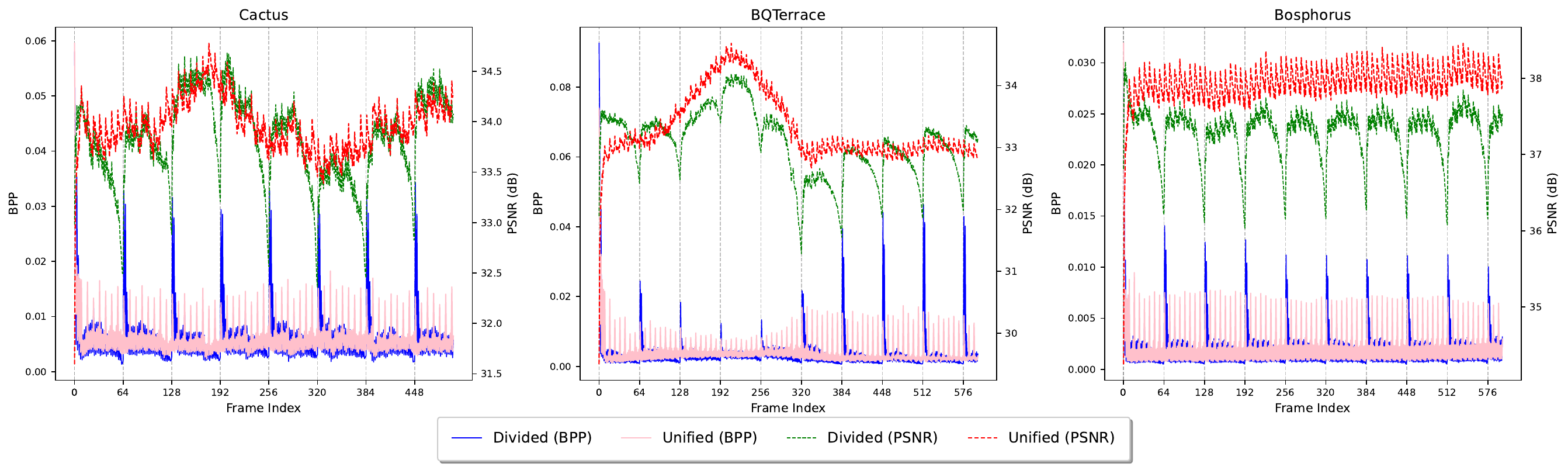}
  \caption{Bitrate and quality variation across frames. The tested videos are Cactus, BQTerrace, and Bosphorus, from left to right. We compare two settings: divided (intra and inter frames use two models, refresh period is 64) and unified (all frames use the same model, no refresh). \emph{Note that we test single-frame compression models in this figure, rather than the proposed two-frame compression scheme.} It is evident that the unified model delivers more stable bitrate and quality per frame and avoids the need for refresh, even in the setting of single-frame compression. Yet, our two-frame compression scheme performs even better. Detailed results for more sequences are provided in the supplementary material.}
  \label{fig:all_seq}
  \vspace{-5mm}
\end{figure*}

\section{Experiments}
\subsection{Setting}
\textbf{Datasets.} We use Vimeo-90k~\cite{xue2019video} to train our model with 7-frame sequences. The original Vimeo videos\footnote{\url{https://github.com/anchen1011/toflow/blob/master/data/original_vimeo_links.txt}} are then cropped into longer sequences for fine-tuning by following \cite{jia2025towards}. We evaluate our model on the test sets HEVC Class B$\sim$E~\cite{flynn16common}, UVG~\cite{uvg}, and MCL-JCV~\cite{mcl-jcv}.

\textbf{Experimental Settings.} For traditional codecs, we compare with VTM, and the detailed parameters are provided in the Appendix. For neural codecs, we compare with state-of-the-art open-source models NVCs, including DCVC-DC~\cite{dc}, DCVC-FM~\cite{li2024neural}, and DCVC-RT~\cite{jia2025towards}. Since our work primarily targets real-time applications, we follow the main testing scenario of RT in the YUV420 color space. All tests are conducted under the low-delay configuration. The bitrate performance is evaluated using the estimated entropy \cite{li2024neural}.
Therefore, our results may slightly differ from those reported in~\cite{jia2025towards}.
Because two frames share a common bitstream in our model, for each frame, we take the average size of the shared bitstream as the bitrate of that frame, which does not affect the overall bitrate computation or BD-rate calculation and can reflect the fluctuation of bitrate.
All experiments are performed on a unified hardware setup: an NVIDIA GeForce RTX 3090 GPU and an Intel(R) Xeon(R) Gold 6248R CPU @ 3.00GHz. We report the average frame rate at different quantization parameters (QP) for the resolution of 1920$\times$1080.
For testing computational complexity, we uniformly used the method for testing complexity provided in the DeepSpeed\footnote {\url{https://github.com/microsoft/DeepSpeed}}  library.

\textbf{Training Details.} To achieve multi-rate, we randomly selected different quantization parameters (QPs) in the range [0, 63] for each training iteration. In a group of 8 images, the QP bias selection was [0, 8, 0, 4, 0, 4, 0, 4] for hierarchical quality. We followed \cite{dc} to assign different hierarchical weights to the loss function of each frame to support the hierarchical quality structure. 
For the loss function, we used a scaled YUV mean squared error loss. The training was conducted on 8 NVIDIA RTX 4090 GPUs.

\subsection{Comparison results}
In Table \ref{tab:results}, we present the BD-rate comparison under the YUV420 format with the intra period set to -1 across the full test sequences. As indicated in the table, our model achieves an average bitrate saving of 35.7\% compared to VTM. Furthermore, it outperforms the state-of-the-art neural video coding (NVC) method DCVC-FM by 6.8\% in rate-distortion performance, while operating at approximately 25$\times$ faster encoding and decoding speeds, reaching 65.1 fps for encoding and 46.1 fps for decoding. Compared to the only practical real-time NVC method, DCVC-RT, our model improves coding performance by 12.1\% under nearly identical encoding and decoding speeds. These results demonstrate the excellent rate-distortion performance and computational efficiency of our proposed model.

Figure \ref{fig:rd-curve} illustrates the rate-distortion curves on the HEVC-E and HEVC-B test sets. Our model consistently outperforms VTM across almost all quality levels. In particular, it shows strong performance at low bitrates. At high bitrates, thanks to reduced error accumulation in long sequences, our model even surpasses the highly complex DCVC-FM on HEVC-E. However, our model does not perform as well on shorter sequences, especially on MCL-JCV, which contains sequences with a maximum length of only 150 frames. Additionally, further exploration of training strategies is still needed---for instance, our reproduced version of DCVC-RT still trails the official model. Nevertheless, under almost similar training settings, our model already outperforms the released version of DCVC-RT.

Moreover, our model exhibits more stable bitrate and quality fluctuations, and recovers more rapidly to higher quality after scene changes, as shown in Figure \ref{fig:kimono}. Evaluated on the kimono sequence, which contains explicit scene cuts, our model restores quality noticeably faster than DCVC-RT, demonstrating a stronger ability to cope with abrupt content transitions.
Moreover, our approach maintains consistently high quality without requiring refresh operations and remains unaffected by error propagation. Moreover, while sustaining higher quality, our method achieves better bitrate savings, and the peak bitrate has also been reduced. Detailed results for more sequences are provided in the supplementary material.

In Figure \ref{fig:rdintra}, we compare the intra-coding performance of our model with that of DCVC-RT. It can be observed that the intra-frame compression ability of our model is significantly better than that of DCVC-RT’s P-frame model, and is only slightly worse than DCVC-RT’s high-complexity I-frame model. 
This experiment suggests that our UI$^2$C’s intrinsic intra-coding capacity is markedly superior to that of DCVC-RT’s P-frame model, allowing it to handle previously unseen scenes more effectively.

\subsection{Complexity analysis}
Table \ref{tab:complexity} provides a complexity comparison among different models. Compared to DCVC-DC and DCVC-FM, our model achieves a better compression ratio while maintaining lower computational complexity. Relative to DCVC-RT, our approach exhibits slightly higher encoding and decoding complexity. However, since our method processes two frames jointly during encoding and decoding, the average latent size per frame and the number of decoding steps are reduced by half. The overall frame rates of our method and DCVC-RT are comparable. It should be noted that our method introduces higher latency due to the one-frame delay in the encoding process. This trade-off is acceptable in scenarios where low latency is not strictly required.

\subsection{Ablation studies}

To validate the effectiveness of each technique employed to enhance the performance of our model, we conduct comprehensive ablation studies. The average BD-Rate, computed in terms of YUV420 PSNR on the HEVC test sequences, is utilized as the evaluation metric.

\textbf{Unified Intra- and Inter-Frame Coding.}
To evaluate the efficacy of unifying intra- and inter-frame coding within a single model, we perform a comparative experiment. Specifically, for the baseline without this technique, we employ the I-frame model from RT to handle intra-frame coding. 
Table ~\ref{tab:ablation}  (Rows 1 vs. 5)  demonstrates that when an I-frame model is available, the refresh mechanism proposed by DCVC-FM leads to significant performance gains. 
However, in the absence of the refresh mechanism, severe error accumulation occurs, resulting in substantially degraded performance. 

Meanwhile, we explored a unified model for both intra-frame and inter-frame processing within a single frame. Experiments show that after enhancing the intra-frame capability, the model's performance improved by 64.9\% under the non-refresh IP-1 condition (Table ~\ref{tab:ablation}, Rows 5 vs. 6). However, under the refresh condition, the performance was inferior to the non-refresh scenario (Table ~\ref{tab:ablation}, Rows 2 vs. 6), as the model tends to lose rich temporal feature information and was not optimized using three-channel images of I-frames during training, while the unified model already handles error propagation effectively. 

Additionally, as shown in the Figure~\ref{fig:all_seq}, after unifying the intra- and inter-frame models, the quality and bitrate of the model became more stable throughout long sequences, with the peak bitrate significantly lower than that of the original refresh method. This makes the model more suitable for practical application scenarios.

\textbf{Simultaneous Two-Frame Compression.}
To assess the effectiveness of the two-frame compression technique, we compare it against the model configuration without any frame delay. As presented in Table~\ref{tab:ablation}, the introduction of the two-frame compression technique yields a remarkable performance improvement, even in scenarios without any refresh operation.

\textbf{Training with Hybrid References.}
To investigate the benefit of the hybrid reference training strategy, we compare it with a baseline that uses only a single blank reference frame. Specifically, during training, we replace the hybrid reference with a single blank frame. Results in Table~\ref{tab:ablation} indicate that our proposed method achieves an RD performance improvement of approximately 5.3\% compared to using a single blank reference.

\section{Conclusion}
This work addresses critical bottlenecks in real-time Neural Video Compression (NVC), including insufficient intra-coding capabilities for inter-frames, error propagation, and bitrate surges. We introduce a unified intra-inter model that, combined with a simultaneous two-frame compression mechanism, effectively tackles these issues.  Experiments demonstrate that our scheme outperforms DCVC-RT by an average of 12.1\% in BD-rate reduction, while maintaining stable bitrate, consistent quality, and comparable real-time inference speed.

Despite these promising results, our model has limitations. Its inference speed is not yet fully optimized for resource-constrained edge devices, such as those with less powerful GPUs or NPUs. Furthermore, its compression efficiency at high bitrates lags behind more complex, non-real-time NVC methods. Future work will focus on developing more lightweight network architectures to reduce computational complexity and integrating advanced modules to boost compression performance at high bitrates.